\documentclass[11pt]{article}
\usepackage{coling2020}
\usepackage{times}
\usepackage{url}
\usepackage{latexsym}

\usepackage{caption}
\usepackage{url}
\usepackage{algorithm}
\usepackage{algorithmic}
\usepackage{graphicx}
\usepackage{amsmath}
\usepackage{amsfonts}
\usepackage{multirow}
\usepackage{booktabs}
\usepackage{xcolor}

\colingfinalcopy 

\title{A Survey of Unsupervised Dependency Parsing}

\author{Wenjuan Han$^1$\footnotemark[1], Yong Jiang$^2$\footnotemark[1], Hwee Tou Ng$^1$, Kewei Tu$^3$\footnotemark[2]\\
$^1$Department of Computer Science, National University of Singapore \\
$^2$Alibaba DAMO Academy, Alibaba Group \\
$^3$School of Information Science and Technology, ShanghaiTech University \\

{\tt dcshanw@nus.edu.sg} \\
{\tt yongjiang.jy@alibaba-inc.com} \\
{\tt nght@comp.nus.edu.sg} \\
{\tt tukw@shanghaitech.edu.cn} \\
\\
}

\date{}

\begin{document}
\maketitle

\renewcommand{\thefootnote}{\fnsymbol{footnote}}
\footnotetext[1]{Equal contributions.}
\footnotetext[2]{Corresponding author.}
\renewcommand{\thefootnote}{\arabic{footnote}}

\begin{abstract}

Syntactic dependency parsing is an important task in natural language processing. Unsupervised dependency parsing aims to learn a dependency parser from sentences that have no annotation of their correct parse trees. Despite its difficulty, unsupervised parsing is an interesting research direction because of its capability of utilizing almost unlimited unannotated text data. It also serves as the basis for other research in low-resource parsing. In this paper, we survey existing approaches to unsupervised dependency parsing, identify two major classes of approaches, and discuss recent trends. We hope that our survey can provide insights for researchers and facilitate future research on this topic.

\end{abstract}

\section{Introduction}
Dependency parsing is an important task in natural language processing that aims to capture syntactic information in sentences in the form of dependency relations between words. It finds applications in semantic parsing, machine translation, relation extraction, and many other tasks. 

Supervised learning is the main technique used to automatically learn a dependency parser from data. It requires the training sentences to be manually annotated with their correct parse trees. Such a training dataset is called a \emph{treebank}. A major challenge faced by supervised learning is that treebanks are not always available for new languages or new domains and building a high-quality treebank is very expensive and time-consuming.

There are multiple research directions that try to learn dependency parsers with few or even no syntactically annotated training sentences, including transfer learning, unsupervised learning, and semi-supervised learning. Among these directions, unsupervised learning of dependency parsers (a.k.a. \emph{unsupervised dependency parsing} and dependency grammar induction) is the most challenging, which aims to 
obtain a dependency parser 
without using annotated sentences.
Despite its difficulty, unsupervised parsing is an interesting research direction, not only because it would reveal ways to utilize almost unlimited text data without the need for human annotation, but also because it can serve as the basis for studies of transfer and semi-supervised learning of parsers.
The techniques developed for unsupervised dependency parsing could also be utilized for other NLP tasks, such as unsupervised discourse parsing \cite{nishida2020unsupervised}.
In addition, research in unsupervised parsing inspires and verifies cognitive research of human language acquisition.

In this paper, we conduct a survey of unsupervised dependency parsing research.
We first introduce the definition and evaluation metrics of unsupervised dependency parsing, and discuss research areas related to it. Then we present in detail two major classes of approaches to unsupervised dependency parsing: generative approaches and discriminative approaches. Finally, we discuss important new techniques and setups of unsupervised dependency parsing that appear in recent years.





\section{Background}
\subsection{Problem Definition}
Dependency parsing aims at discovering the syntactic dependency tree $\mathbf{z}$ of an input sentence $\mathbf{x}$, where $\mathbf{x}$ is a sequence of words $x_1,\dots,x_n$ with length $n$. A dummy root word $x_0$ is typically added at the beginning of the sentence. A dependency tree $\mathbf{z}$ is a set of directed edges between words that form a directed tree structure rooted at $x_0$. Each edge points from a parent word (also called a head word) to a child word. 

In unsupervised dependency parsing, the goal is to obtain a dependency parser without using annotated sentences. Some work requires no training data and derives dependency trees from centrality or saliency information \cite{sogaard2012unsupervised}. We focus on learning a dependency parser from an unannotated dataset that consists of a set of sentences without any parse tree annotation.
In many cases, part-of-speech (POS) tags of the words in the training sentences are assumed to be available during training. 

Two evaluation metrics are widely used in previous work of unsupervised dependency parsing 
\cite{Klein:2004:CIS:1218955.1219016}: 
directed dependency accuracy (DDA) and undirected dependency accuracy (UDA). DDA denotes the percentage of correctly predicted dependency edges, while UDA is similar to DDA but disregards the directions of edges when evaluating their correctness.

\subsection{Related Areas}
\paragraph{Supervised Dependency Parsing}
Supervised dependency parsing aims to train a dependency parser from training sentences that are manually annotated with their dependency parse trees. Generally, supervised dependency parsing approaches can be divided into graph-based approaches and transition-based approaches. 
A graph-based dependency parser searches for the best spanning tree of the graph that is formed by connecting all pairs of words in the input sentence. In the simplest form, a graph-based parser makes the first-order assumption that the score of a dependency tree is the summation of scores of its edges \cite{mcdonald2005non}. 
A transition-based dependency parser searches for a sequence of actions that incrementally constructs the parse tree, typically from left to right. 
While current start-of-the-art approaches have achieved strong results in supervised dependency parsing, their usefulness is limited to resource-rich languages and domains with many annotated datasets.

\paragraph{Cross-Domain and Cross-Lingual Parsing}
One useful approach to handling the lack of treebank resources in the target domain or language is to adapt a learned parser from a resource-rich source domain or language \cite{yu2015domain,mcdonald2011multi,ma2014unsupervised,duong2015cross}. This is very related to unsupervised parsing as both approaches do not rely on treebanks in the target domain or language. However, unsupervised parsing is more challenging because it does not have access to any source treebank either.

\paragraph{Unsupervised Constituency Parsing} 
Constituency parsing aims to discover a constituency tree of the input sentence in which the leaf nodes are words and the non-leaf nodes (nonterminal nodes) represent phrases. Unsupervised constituency parsing is often considered more difficult than unsupervised dependency parsing because it has to induce not only edges but also nodes of a tree. Consequently, there have been far more papers in unsupervised dependency parsing than in unsupervised constituency parsing over the past decade. More recently, however, there is a surge in interest in unsupervised constituency parsing and several novel approaches were proposed in the past two years \cite{li-etal-2020-empirical}.
While we focus on unsupervised dependency parsing in this paper, most of our discussions on the classification of approaches and recent trends apply to unsupervised constituency parsing as well.

\paragraph{Latent Tree Models with Downstream Tasks} 
Latent tree models treat the parse tree as a latent variable that is used in downstream tasks such as sentiment classification. While no treebank is used in training, these models rely on the performance of the downstream tasks to guide the learning of the latent parse trees.
To enable end-to-end learning, the REINFORCE algorithm and the Gumbel-softmax trick \cite{jang2016categorical} can be utilized \cite{yogatama2016learning,choi2017unsupervised}. There also exists previous work on latent dependency tree models that utilizes structured attention mechanisms \cite{kim2017structured} for applications.
Latent tree models differ from unsupervised parsing in that they utilize training signals from downstream tasks and that they aim to improve performance of downstream tasks instead of syntactic parsing.

\section{General Approaches} \label{sec:approaches}
\subsection{Generative Approaches}\label{sec:generative}
\subsubsection{Models}
A generative approach models the joint probability of the sentence and the corresponding parse tree. Traditional generative models are mostly based on probabilistic grammars. To enable efficient inference, they typically make one or more relatively strict conditional independence assumptions. The simplest assumption (a.k.a. the context-free assumption) states that the generation of a token is only dependent on its head token and is independent of anything else. Such assumptions make it possible to decompose the joint probability into a product of component probabilities or scores, leading to tractable inference. However, they also lead to unavailability of useful information (e.g., context and generation history) in generating each token.

Based on their respective independence assumptions, different generative models specify different generation processes of the sentence and parse tree. \newcite{paskin2002grammatical} and \newcite{carroll1992two} choose to first uniformly sample a dependency tree skeleton and then populate the tokens (words) conditioned on the dependency tree in a recursive root-to-leaf manner. The generation of a child token is conditioned on the head token and the dependency direction.
In contrast, \newcite{Klein:2004:CIS:1218955.1219016} propose the Dependency Model with Valence (DMV) that generates the sentence and the parse tree simultaneously. Without knowing the dependency tree structure, each head token has to sample a decision (conditioned on the head token and the dependency direction) of whether to generate a child token or not before actually generating the child token. Besides, the generation of a child token in DMV is additionally conditioned on the valence, defined as the number of the child tokens already generated from a head token.
\newcite{headden2009improving} propose to also introduce the valence into the condition of decision sampling.
\newcite{spitkovsky2012three} additionally condition decision and child token generation on sibling words, sentence completeness, and punctuation context.
In addition to these generative dependency models, other grammar formalisms have also been used for unsupervised dependency parsing, such as tree substitution grammars \cite{blunsom2010unsupervised} and combinatory categorial grammars \cite{bisk2012simple,bisk2013hdp}.



Similar tokens may have similar syntactic behaviors in a grammar. For example, all the verbs are very likely to generate a noun to the left as the subject. One way to capture this prior knowledge is to compute generation probabilities from a set of features that conveys syntactic similarity.
\newcite{berg2010painless} use a log-linear model based on manually-designed local morpho-syntactic features (e.g., whether a word is a noun) and \newcite{jiang-han-tu:2016:EMNLP2016} employ a neural network to automatically learn such features. Both approaches are based on DMV.

\subsubsection{Inference}
Given a model parameterized by $\Theta$ and a sentence $\mathbf{x}$, the model predicts the parse $\mathbf{z}^{*}$ with the highest probability.
\begin{equation}
\begin{split}
\mathbf{z^{*}} = \arg\max_{\mathbf{z} \in  \mathcal{Z}(\mathbf{x})} P(\mathbf{x}, \mathbf{z};\Theta)
\end{split}
\end{equation}
where $\mathcal{Z}(\mathbf{x})$ is the set of all valid dependency trees of the sentence $\mathbf{x}$.
Due to the independence assumptions made by generative models, the inference problem can be efficiently solved exactly in most cases. For example, chart parsing can be used for DMV.

\subsubsection{Learning Objective}
Log marginal likelihood is typically employed as the objective function for learning generative models. 
It is defined on $N$ training sentences $X=\{\mathbf{x}^{(1)}, \mathbf{x}^{(2)},..., \mathbf{x}^{(N)}\}$:
\begin{equation}\label{obj}
\begin{split}
L(\Theta) =\mathop{\sum_{i = 1}^{N} \log \mathrm{P}(\mathbf{x}^{(i)}; \Theta)}
\end{split}
\end{equation}
where the model parameters are denoted by $\Theta$. The likelihood of each sentence $\mathbf{x}$ is as follows:
\begin{equation}
\begin{split}
P(\mathbf{x};\Theta) = \sum_{\mathbf{z} \in  \mathcal{Z}(\mathbf{x})} P(\mathbf{x}, \mathbf{z};\Theta)
\end{split}
\end{equation}
where $\mathcal{Z}(\mathbf{x})$ is the set of all valid dependency trees of sentence $\mathbf{x}$. As we mentioned earlier, the joint probability of a sentence and its dependency tree can be decomposed into the product of the probabilities of the components in the dependency tree.

Apart from the vanilla marginal likelihood, priors and regularization terms are often added into the objective function to incorporate various inductive biases.
\newcite{smith2006annealing} insert penalty terms into the objective to control dependency lengths and the root number of the parse tree.
Cohen and Smith \shortcite{cohen2008logistic,cohen2009shared} leverage logistic-normal prior distributions to encourage correlations between POS tags in DMV. 
\newcite{naseem2010using} design a posterior constraint based on a set of manually-specified universal dependency rules.
\newcite{gillenwater2011posterior} add a posterior regularization term to encourage rule sparsity.
The approaches of \newcite{spitkovsky2011punctuation} can be seen as adding posterior constraints over parse trees based on punctuation.
\newcite{tu2012unambiguity} introduce an entropy term to prevent the model from becoming too ambiguous.
\newcite{marevcek2012exploiting} insert a term that prefers reducible subtrees (i.e., their removal does not break the grammaticality of the sentence) in the parse tree.
The same reducibility principle is used by \newcite{marecek2013stop} to bias the decision probabilities in DMV.
\newcite{noji2016using} place a hard constraint in the objective that limits the degree of center-embedding of the parse tree.

\subsubsection{Learning Algorithm}
The Expectation-Maximization (EM) algorithm is typically used to optimize log marginal likelihood. For each sentence, the EM algorithm aims to maximize the following lower-bound of the objective function and alternates between the E-step and M-step. 
\begin{equation}\label{obj_lowerBound}
\begin{split}
\log P(\mathbf{x};\Theta) 
- KL(Q(\mathbf{z})\|P(\mathbf{z}|\mathbf{x},\Theta))
\end{split}
\end{equation}
where $Q(\mathbf{z})$ is an auxiliary distribution with regard to $\mathbf{z}$. In the E-step, $\Theta$ is fixed and $Q(\mathbf{z})$ is set to $P(\mathbf{z}|\mathbf{x},\Theta)$. A set of so-called expected counts can be derived from $Q(\mathbf{z})$ to facilitate the subsequent M-step and they are typically calculated using the inside-outside algorithm. In the M-step, $\Theta$ is optimized based on the expected counts with $Q(\mathbf{z})$ fixed.

There are a few variants of the EM algorithm. If $Q(\mathbf{z})$ represents a point-estimation (i.e., the best dependency tree has a probability of 1), the algorithm becomes hard-EM or Viterbi EM, which is found to outperform standard EM in unsupervised dependency parsing \cite{spitkovsky2010viterbi}. Softmax-EM \cite{tu2012unambiguity} falls between EM (considering all possible dependency trees) and hard-EM (only considering the best dependency tree), applying a softmax-like transformation to $Q(\mathbf{z})$. 
During the EM iterations, an annealing schedule \cite{tu2012unambiguity} can be used to gradually shift from hard-EM to softmax-EM and finally to the EM algorithm, which leads to better performance than sticking to a single algorithm. Lateen EM \cite{spitkovsky2011lateen} repeatedly alternates between EM and hard-EM, which is also found to produce better results than both EM and hard-EM.

Approaches with more complicated objectives often require more advanced learning algorithms, but many of the algorithms can still be seen as extensions of the EM algorithm that revise either the E-step (e.g., to update $Q(\mathbf{z})$ based on posterior regularization terms) or the M-step (e.g., to optimize the posterior probability that incorporates parameter priors).


Better learning results can also be achieved by manipulating the training data. \newcite{SpitkovskyEtAl10} apply curriculum learning to DMV training, which starts with only the shortest sentences and then progresses to increasingly longer sentences. 
\newcite{tu2011utility} provide a theoretical analysis on the utility of curriculum learning in unsupervised dependency parsing.

\newcite{spitkovsky2013breaking} propose to treat different learning algorithms and configurations as modules and connect them to form a network. Some approaches discussed above, such as Lateen EM and curriculum learning, can be seen as special cases of this approach.


\subsubsection{Pros and Cons}
It is often straightforward to incorporate various inductive biases and manually-designed local features into generative approaches. Moreover, generative models can be easily trained via the EM algorithm and its extensions. On the other hand, generative models often have limited expressive power because of the independence assumptions they make.

\subsection{Discriminative Approaches}
Because of the limitation of generative approaches, more recently, researchers have paid more attention to discriminative approaches. Discriminative approaches model the conditional probability or score of the dependency tree given the sentence. By conditioning on the whole sentence, discriminative approaches are capable of utilizing not only local features (i.e., features related to the current dependency) but also global features (i.e., contextual features from the whole sentence) in scoring a dependency tree.

\begin{table}[]
\centering
\resizebox{0.7\textwidth}{!}{%
\begin{tabular}{cc|ccc}
\hline
 &  & \textbf{\begin{tabular}[c]{@{}c@{}}Intermediate\\ Representation\end{tabular}} & \textbf{Encoder} & \textbf{Decoder} \\ \hline
\multicolumn{1}{c|}{\multirow{2}{*}{Autoencoder}} & CRFAE \cite{cai2017crf} & Z & $P(\mathbf{z}|\mathbf{x})$ & $P(\mathbf{\hat{x}}|\mathbf{z})$ \\ \cline{2-5} 
\multicolumn{1}{c|}{} & \begin{tabular}[c]{@{}c@{}} D-NDMV \cite{han2019enhancing}\\ Deterministic Variant \end{tabular} & S & $P(\mathbf{s}|\mathbf{x})$ &  $P(\mathbf{z}, \mathbf{\hat{x}}|\mathbf{s})$\\ \hline
\multicolumn{1}{c|}{\multirow{3}{*}{\begin{tabular}[c]{@{}c@{}}Variational\\ Autoencoder\end{tabular}}} & \cite{li2019dependency}  & Z & $P(\mathbf{z}|\mathbf{x})$ & $P(\mathbf{z},\mathbf{x})$ \\ \cline{2-5} 
\multicolumn{1}{c|}{} & \begin{tabular}[c]{@{}c@{}}D-NDMV \cite{han2019enhancing}\\ Variational Variant \end{tabular} & S & $P(\mathbf{s}|\mathbf{x})$ & $P(\mathbf{z},\mathbf{x}|\mathbf{s})$ \\ \cline{2-5} 
\multicolumn{1}{c|}{} & \cite{corro2018differentiable} & Z & $P(\mathbf{z}|\mathbf{x})$ & $P(\mathbf{x}|\mathbf{z})$ \\ \hline
\end{tabular}%
}
\caption{Major approaches based on autoencoders and variational autoencoders for unsupervised dependency parsing. Z: dependency tree. S: continuous sentence representation. $\mathbf{\hat{x}}$ is a copy of $\mathbf{x}$ representing the reconstructed sentence. $\mathbf{z}$ is the dependency tree. $\mathbf{s}$ is the continuous representation of sentence $\mathbf{x}$.}
\label{table:ae_vae}
\end{table}

\subsubsection{Autoencoder-Based Approaches}\label{sec:ae}
Autoencoder-based approaches aim to map a sentence to an intermediate representation (encoding) and then reconstruct the observed sentence from the intermediate representation (decoding). In the two existing autoencoder approaches (summarized in Table \ref{table:ae_vae}), the intermediate representation is the dependency tree and a continuous sentence vector respectively.


The reconstruction loss is typically employed as the learning objective function for autoencoder models. For a training dataset including $N$ sentences $X=\{\mathbf{x}^{1}, \mathbf{x}^{2},..., \mathbf{x}^{N}\}$, the objective function is as follows: 
\begin{equation}\label{autodencoder:obj}
\begin{split}
L(\Theta) =\mathop{\sum_{i = 1}^{N} \log \mathrm{P}(\mathbf{\hat{x}}^{(i)}|\mathbf{x}^{(i)}; \Theta)}
\end{split}
\end{equation}
where $\Theta$ is the model parameter and $\mathbf{\hat{x}}^{(i)}$ is a copy of $\mathbf{x}^{(i)}$ representing the reconstructed sentence\footnote{In \newcite{han2019enhancing}, $\mathbf{x}$ is the word sequence, while $\mathbf{\hat{x}}$ is the POS tag sequence of the same sentence.}. In some cases, there is an additional regularization term (e.g., L1) of $\Theta$. 

The first autoencoder model for unsupervised dependency parsing,  proposed by \newcite{cai2017crf}, is based on the conditional random field autoencoder framework (CRFAE). The encoder is a first-order graph-based discriminative dependency parser mapping an input sentence to the space of dependency trees. The decoder independently generates each token of the reconstructed sentence conditioned on the head of the token specified by the dependency tree. Both the encoder and the decoder are arc-factored, meaning that the encoding and decoding probabilities can be factorized by dependency arcs. 
Coordinate descent is applied to minimize the reconstruction loss and alternately updates the encoder parameters and the decoder parameters.

D-NDMV \cite{han2019enhancing} (the deterministic variant) is the second autoencoder model proposed for unsupervised dependency parsing, in which the intermediate representation is a continuous vector representing the input sentence. The encoder is an LSTM summarizing the sentence with a continuous vector $\mathbf{s}$, while the decoder models the joint probability of the sentence and the dependency tree. More specifically, the decoder is a generative neural DMV that generates the sentence and its parse simultaneously, and its parameters are computed based on the continuous vector $\mathbf{s}$. The reconstruction loss is optimized using the EM algorithm. In the E-step, $\Theta$ is fixed and $Q(\mathbf{z})$ is set to $P(\mathbf{z}|\mathbf{x},\mathbf{s};\Theta)$. After we compute all the grammar rule probabilities given $\Theta$, the inside-outside algorithm can be used to calculate the expected counts. In the M-step, $\Theta$ is optimized based on the expected counts with $Q(\mathbf{z})$ fixed.

\subsubsection{Variational Autoencoder-Based Approaches}\label{sec:vae}
As mentioned in Section \ref{sec:generative}, the training objective of a generative model is typically the probability of the training sentence and the dependency tree is marginalized as a hidden variable. However, the marginalized probability cannot usually be calculated accurately for more complex models that do not make strict independence assumption. Instead, a variational autoencoder maximizes the Evidence Lower Bound (ELBO), a lower bound of the marginalized probability. Since the intermediate representation follows a distribution, different sampling approaches are used to optimize the objective function (i.e., likelihood) according to different model schema. 

Three unsupervised dependency parsing models were proposed in recent years based on variational autoencoders (shown in Table \ref{table:ae_vae}).
There are three probabilities involved in ELBO: the prior probability of the syntactic structure, the probability of generating the sentence from the syntactic structure (the decoder), and the variational posterior (the encoder) from the sentence to the syntactic structure.

Recurrent Neural Network Grammars (RNNG) \cite{dyer2016recurrent} is a transition-based constituent parser, with a discriminative and a generative variant. Discriminative RNNG incrementally constructs the constituency tree of the input sentence through three kinds of operations: generating a non-terminal token, shifting, and reducing. Generative RNNG replaces the shifting operation with a word generation operation and incrementally generates a constituency tree and its corresponding sentence. The probability of each operation is calculated by a neural network. \newcite{li2019dependency} modify RNNG for dependency parsing and use discriminative RNNG and generative RNNG as the encoder and decoder of a variational autoencoder respectively. However, because RNNG has a strong expressive power, it is prone to overfitting in the unsupervised setting. \newcite{li2019dependency} propose to use posterior regularization to introduce linguistic knowledge as a constraint in learning, thereby mitigating this problem to a certain extent.

The model proposed by \newcite{corro2018differentiable} is also based on a variational autoencoder. It is designed for semi-supervised dependency parsing, but in principle it can also be applied for unsupervised dependency parsing. The encoder of this model is a conditional random field model while the decoder generates a sentence based on a graph convolutional neural network whose structure is specified by the dependency tree. Since the variational autoencoder needs Monte Carlo sampling to approximate the gradient and the complexity of sampling a dependency tree is very high, \newcite{corro2018differentiable} use Gumbel random perturbation. \newcite{jang2016categorical} use differentiable dynamic programming to design an efficient approximate sampling algorithm.

The variational variant of D-NDMV \cite{han2019enhancing} has the same structure as the deterministic variant described in Section \ref{sec:ae}, except that the variational variant probabilistically models the intermediate continuous vector conditioned on the input sentence using a Gaussian distribution. It also specifies a Gaussian prior over the intermediate continuous vector.

\subsubsection{Other Discriminative Approaches}
Apart from the approaches based on autoencoder and variational autoencoder, there are also a few other discriminative approaches based on discriminative clustering \cite{grave2015convex}, self-training \cite{le2015unsupervised}, or searching \cite{daume2009unsupervised}.
Because of space limit, below we only introduce the approach based on discriminative clustering called Convex MST \cite{grave2015convex}.

Convex MST employs a first-order graph-based discriminative parser. It searches for the parses of all the training sentences and learns the parser simultaneously, with a learning objective that the searched parses are close to the predicted parses by the parser. In other words, the parses should be easily predictable by the parser. The objective function can be relaxed to become convex and then can be optimized exactly.

\subsubsection{Pros and Cons}

Discriminative models are capable of accessing global features from the whole input sentence and are typically more expressive than generative models. On the other hand, discriminative approaches are often more complicated and do not admit tractable exact inference.













\section{Recent Trends} \label{sec:trends}
\subsection{Combined Approaches} 
Generative approaches and discriminative approaches have different pros and cons. Therefore, a natural idea is to combine the strengths of the two types of approaches to achieve better performance. \newcite{jiang-etal-2017-combining} propose to jointly train two state-of-the-art models of unsupervised dependency parsing, the generative LC-DMV \cite{noji2016using} and the discriminative Convex MST, with the dual decomposition technique  that encourages the
two models to gradually influence each other during training.

\subsection{Neural Parameterization}
Traditional generative approaches either directly learn or use manually-designed features to compute dependency rule probabilities. Following the recent rise of deep learning in the field of NLP, \newcite{jiang-han-tu:2016:EMNLP2016} propose to predict dependency rule probabilities using a neural network that takes as input the vector representations of the rule components such as the head and child tokens. The neural network can automatically learn features that capture correlations between tokens and rules. \newcite{han2019enhancing} extend this generative approach to a discriminative approach by further introducing sentence information into the neural network in order to compute sentence-specific rule probabilities. Compared with generative approaches, it is more natural for discriminative approaches to use neural networks to score dependencies or parsing actions, so recent discriminative approaches all make use of neural networks \cite{li2019dependency,corro2018differentiable}.

\subsection{Lexicalization}
In the most common setting of unsupervised dependency parsing, the parser is unlexicalized with POS tags being the tokens in the sentences. The POS tags are either human annotated or induced from the training corpus \cite{spitkovsky2011unsupervised,he2018unsupervised}.
However, words with the same POS tag may have very different syntactic behavior and hence it should be beneficial to introduce lexical information into unsupervised parsers.
\newcite{headden2009improving}, \newcite{blunsom2010unsupervised}, and \newcite{han2017dependency} use partial lexicalization in which infrequent words are replaced by special symbols or their POS tags.  \newcite{pategrammar} and \newcite{spitkovsky2013breaking} experiment with full lexicalization. However, because the number of words is huge, a major problem with full lexicalization is that the grammar becomes much larger and thus learning requires more data. To mitigate the negative impact of data scarcity, smoothing techniques can be used. For instance, \newcite{han2017dependency} use neural networks to predict dependency probabilities that are automatically smoothed. 

In principle, lexicalized approaches could also benefit from pretrained word embeddings, which capture syntactic and semantic similarities between words. Recently proposed contextual word embeddings \cite{devlin2018bert} are even more informative, capturing contextual information. However, word embeddings have not been widely used in unsupervised dependency parsing. One concern is that word embeddings are too informative and may make unsupervised models more prone to overfitting.
One exception is \newcite{he2018unsupervised}, who propose to use invertible neural projections to map word embeddings into a latent space that is more amenable to unsupervised parsing.

\subsection{Big Data}
Although unsupervised parsing does not require syntactically annotated training corpora and can theoretically use almost unlimited raw texts for training, most of the previous work conducts experiments on the WSJ10 corpus (the Wall Street Journal corpus with sentences no longer than 10 words) containing no more than 6,000 training sentences. There are a few papers that try to go beyond such a small training corpus. \newcite{pategrammar} use two large corpora containing more than 700k sentences. \newcite{marecek2013stop} utilize a very large corpus based on Wikipedia in learning an unlexicalized dependency grammar. \newcite{han2017dependency} use a subset of the BLLIP corpus that contains around 180k sentences. With the advancement of computing power and deep neural models, we expect to see more future work on training with big data.

\subsection{Unsupervised Multilingual Parsing}
To tackle the lack of supervision in unsupervised dependency parsing, some previous work considers learning models of multiple languages simultaneously \cite{berg2010phylogenetic,liu2013bilingually,jiang-etal-2019-regularization,han-etal-2019-multilingual}. Ideally, these models can learn from each other by identifying shared syntactic behaviors of different languages, especially those in the same language family. For example, \newcite{berg2010phylogenetic} propose to utilize the similarity of different languages defined by a phylogenetic tree and learn several dependency parsers jointly. \newcite{han-etal-2019-multilingual} propose to learn a unified multilingual parser with language embeddings as input. \newcite{jiang-etal-2019-regularization} propose to guide the learning process of unsupervised dependency parser from the knowledge of another language by using three types of regularization to encourage similarity between model parameters, dependency edge scores, and parse trees respectively. 

\begin{table}[]
\centering
\small
\resizebox{0.9\textwidth}{!}{%
\begin{tabular}{c|c|c||c|c|c}
\hline \hline
{\bf \textsc{Methods}}                             & {\bf \textsc{$\leq10$}} & {\bf\textsc{All}}  & \multicolumn{3}{c}{Generative Approaches (cont'd)}                                                                                                                                                                \\ \cline{1-3} \hline 
\multicolumn{3}{c||}{Generative Approaches}                                                                                                                                                             & \multicolumn{1}{c|}{ \newcite{spitkovsky2011unsupervised}}  & ~~~-                      & 59.1                                     \\ \cline{1-3}
 \newcite{Klein:2004:CIS:1218955.1219016} & 46.2                                                      & 34.9                                                & \multicolumn{1}{c|}{ \newcite{gimpel2012concavity}}         & \multicolumn{1}{c|}{64.3} & 53.1 \\
 \newcite{cohen2008logistic}              & 59.4                                                      & 40.5                                                & \multicolumn{1}{c|}{ \newcite{tu2012unambiguity}}           & \multicolumn{1}{c|}{71.4}                                     & 57.0                                     \\
  \newcite{cohen2009shared}               & 61.3                                                      & 41.4                                                & \multicolumn{1}{c|}{ \newcite{bisk2012simple}}              & \multicolumn{1}{c|}{71.5}                                     & 53.3                                     \\
 \newcite{headden2009improving}           & 68.8                                                      & ~~~-                                 & \multicolumn{1}{c|}{ \newcite{spitkovsky2013breaking}}      & \multicolumn{1}{c|}{72.0}                                     & 64.4                                     \\
 \newcite{SpitkovskyEtAl10}               & 56.2                                                      & 44.1                                                & \multicolumn{1}{c|}{ \newcite{jiang-han-tu:2016:EMNLP2016}} & \multicolumn{1}{c|}{72.5}                                     & 57.6                                     \\
  \newcite{berg2010painless}              & 63.0                                                      & ~~~-                                 & \multicolumn{1}{c|}{ \newcite{han2017dependency}}           & \multicolumn{1}{c|}{75.1}                                     & 59.5                                     \\
 \newcite{gillenwater2010sparsity}        & 64.3                                                      & 53.3                                                & \multicolumn{1}{c|}{ \newcite{he2018unsupervised}*}         & \multicolumn{1}{c|}{60.2}                                     & 47.9                                     \\ \cline{4-6} 
 \newcite{spitkovsky2010viterbi}          & 65.3                                                      & 47.9                                                & \multicolumn{3}{c}{Discriminative Approaches}                                                                                                                                                                     \\ \cline{4-6} 
 \newcite{blunsom2010unsupervised}        & 65.9                                                      & 53.1                                                & \multicolumn{1}{c|}{\newcite{daume2009unsupervised}}       & ~~~-                      & 45.4                                     \\
 \newcite{naseem2010using}                & 71.9                                                      & ~~~-                                 & \multicolumn{1}{c|}{\newcite{le2015unsupervised} $\dag$}   & \multicolumn{1}{c|}{73.2}                                     & 65.8                                     \\
\newcite{blunsom2010unsupervised}        & 67.7                                                      & 55.7                                                & \multicolumn{1}{c|}{\newcite{cai2017crf}}                  & \multicolumn{1}{c|}{71.7}                                     & 55.7                                     \\
\newcite{spitkovsky2011lateen}           & ~~~-                                       & 55.6                                                & \multicolumn{1}{c|}{\newcite{li2019dependency}}            & \multicolumn{1}{c|}{54.7}                                     & 37.8                                     \\
\newcite{spitkovsky2011punctuation}      & 69.5                                                      & 58.4                                                & \multicolumn{1}{c|}{\newcite{han2019enhancing}}            & \multicolumn{1}{c|}{75.6}                                     & 61.4                                     \\ \hline
\end{tabular}%
}
  \caption{Reported directed dependency accuracies on section 23 of the WSJ corpus, evaluated on sentences of length $\leq10$ and all lengths. *: without gold POS tags. $\dag$: with more training data in addition to WSJ.}
  \label{table:compareRecurrentResults}
\end{table}

\section{Benchmarking on the WSJ Corpus}
Most papers of unsupervised dependency parsing report the accuracy of their approaches on the test set of the Wall Street Journal (WSJ) corpus. We list the reported accuracy on WSJ in Table \ref{table:compareRecurrentResults}. 
It must be emphasized that the approaches listed in this table may use different training sets and different external knowledge in their experiments, and one should check the corresponding papers to understand such differences before comparing these accuracies.

While the accuracy of unsupervised dependency parsing has increased by over thirty points in the last fifteen years, it is still well below that of supervised models, which leaves much room for improvement and challenges for future research.

\section{Future Directions}
\subsection{Utilization of Syntactic Information in Pretrained Language Modeling}
Pretrained language modeling \cite{Peters:2018,devlin2018bert,radford2019language}, as a new NLP paradigm, has been utilized in various areas including question answering, machine translation, grammatical error correction, and so on. Pretrained language models leverage a large-scale corpus for pretraining and then small data sets of specific tasks for finetuning, reducing the difficulty of downstream tasks and boosting their performance. Current state-of-the-art approaches on supervised dependency parsing, such as \newcite{zhou2019head}, adopt the new paradigm and benefit from pretrained language modeling.  
However, pretrained language models have not been widely used in unsupervised dependency parsing. 
One major concern is that pretrained language models are too informative and may make unsupervised models more prone to overfitting. Besides, massive syntactic and semantic information is encoded in pretrained language models and how to extract the syntactic part from them is a challenging task.


\subsection{Inspiration for Other Tasks}
Unsupervised dependency parsing is a classic unsupervised learning task. Many techniques developed for unsupervised dependency parsing can serve as the inspiration for studies of other unsupervised tasks, especially unsupervised structured prediction tasks. A recent example is \newcite{nishida2020unsupervised}, who study unsupervised discourse parsing (inducing discourse structures for a given text) by borrowing techniques from unsupervised parsing such as Viterbi EM and heuristically designed initialization. 


Unsupervised dependency parsing techniques can also be used as building blocks for transfer learning of parsers. Some of the approaches discussed in this paper have already been applied to cross-lingual parsing \cite{he2019cross}, and more such endeavors are expected in the future.

\subsection{Interpretability}

One prominent problem of deep neural networks is that they act as black boxes and are generally not interpretable. How to improve the interpretability of neural networks is a research topic that gains much attention recently. 
For natural language texts, their linguistic structures reveal important information of the texts and at the same time can be easily understood by human. It is therefore an interesting direction to integrate techniques of unsupervised parsing into various neural models of NLP tasks, such that the neural models can build their task-specific predictions on intermediate linguistic structures of the input text, which improves the interpretability of the predictions.

  
\section{Conclusion}
In this paper, we present a survey on the current advances of unsupervised dependency parsing. We first motivate the importance of the unsupervised dependency parsing task and discuss several related research areas. We split existing approaches into two main categories, and explain each category in detail. Besides, we discuss several recent trends in this research area. 
While there is a growing body of work that improves unsupervised dependency parsing, its performance is still below that of supervised dependency parsing by a large margin. This suggests that more investigation and research are needed to make unsupervised parsers useful for real applications. We hope that our survey can promote further development in this research direction.

\section*{Acknowledgments}
Kewei Tu was supported by the National Natural Science Foundation of China (61976139).

\bibliographystyle{coling}
\bibliography{coling2020}

\begin{thebibliography}{}

\bibitem[\protect\citename{Berg-Kirkpatrick and
  Klein}2010]{berg2010phylogenetic}
Taylor Berg-Kirkpatrick and Dan Klein.
\newblock 2010.
\newblock Phylogenetic grammar induction.
\newblock In {\em ACL}.

\bibitem[\protect\citename{Berg-Kirkpatrick \bgroup et al.\egroup
  }2010]{berg2010painless}
Taylor Berg-Kirkpatrick, Alexandre Bouchard-C{\^o}t{\'e}, John DeNero, and Dan
  Klein.
\newblock 2010.
\newblock Painless unsupervised learning with features.
\newblock In {\em NAACL}.

\bibitem[\protect\citename{Bisk and Hockenmaier}2012]{bisk2012simple}
Yonatan Bisk and Julia Hockenmaier.
\newblock 2012.
\newblock Simple robust grammar induction with combinatory categorial grammars.
\newblock In {\em AAAI}.

\bibitem[\protect\citename{Bisk and Hockenmaier}2013]{bisk2013hdp}
Yonatan Bisk and Julia Hockenmaier.
\newblock 2013.
\newblock An {HDP} model for inducing combinatory categorial grammars.
\newblock {\em TACL}.

\bibitem[\protect\citename{Blunsom and Cohn}2010]{blunsom2010unsupervised}
Phil Blunsom and Trevor Cohn.
\newblock 2010.
\newblock Unsupervised induction of tree substitution grammars for dependency
  parsing.
\newblock In {\em EMNLP}.

\bibitem[\protect\citename{Cai \bgroup et al.\egroup }2017]{cai2017crf}
Jiong Cai, Yong Jiang, and Kewei Tu.
\newblock 2017.
\newblock {CRF} autoencoder for unsupervised dependency parsing.
\newblock In {\em EMNLP}.

\bibitem[\protect\citename{Carroll and Charniak}1992]{carroll1992two}
Glenn Carroll and Eugene Charniak.
\newblock 1992.
\newblock Two experiments on learning probabilistic dependency grammars from
  corpora.
\newblock Technical report, Department of Computer Science, Brown University.

\bibitem[\protect\citename{Choi \bgroup et al.\egroup
  }2018]{choi2017unsupervised}
Jihun Choi, Kang~Min Yoo, and Sang-goo Lee.
\newblock 2018.
\newblock Unsupervised learning of task-specific tree structures with
  tree-lstms.
\newblock In {\em AAAI}.

\bibitem[\protect\citename{Cohen and Smith}2009]{cohen2009shared}
Shay~B Cohen and Noah~A Smith.
\newblock 2009.
\newblock Shared logistic normal distributions for soft parameter tying in
  unsupervised grammar induction.
\newblock In {\em NAACL}.

\bibitem[\protect\citename{Cohen \bgroup et al.\egroup
  }2008]{cohen2008logistic}
Shay~B Cohen, Kevin Gimpel, and Noah~A Smith.
\newblock 2008.
\newblock Logistic normal priors for unsupervised probabilistic grammar
  induction.
\newblock In {\em NIPS}.

\bibitem[\protect\citename{Corro and Titov}2018]{corro2018differentiable}
Caio Corro and Ivan Titov.
\newblock 2018.
\newblock Differentiable perturb-and-parse: Semi-supervised parsing with a
  structured variational autoencoder.
\newblock In {\em ICLR}.

\bibitem[\protect\citename{Daum{\'e}~III}2009]{daume2009unsupervised}
Hal Daum{\'e}~III.
\newblock 2009.
\newblock Unsupervised search-based structured prediction.
\newblock In {\em ICML}.

\bibitem[\protect\citename{Devlin \bgroup et al.\egroup }2019]{devlin2018bert}
Jacob Devlin, Ming-Wei Chang, Kenton Lee, and Kristina Toutanova.
\newblock 2019.
\newblock {BERT}: Pre-training of deep bidirectional transformers for language
  understanding.
\newblock In {\em NAACL}.

\bibitem[\protect\citename{Duong \bgroup et al.\egroup }2015]{duong2015cross}
Long Duong, Trevor Cohn, Steven Bird, and Paul Cook.
\newblock 2015.
\newblock Cross-lingual transfer for unsupervised dependency parsing without
  parallel data.
\newblock In {\em CoNLL}.

\bibitem[\protect\citename{Dyer \bgroup et al.\egroup }2016]{dyer2016recurrent}
Chris Dyer, Adhiguna Kuncoro, Miguel Ballesteros, and Noah~A Smith.
\newblock 2016.
\newblock Recurrent neural network grammars.
\newblock In {\em NAACL}.

\bibitem[\protect\citename{Gillenwater \bgroup et al.\egroup
  }2010]{gillenwater2010sparsity}
Jennifer Gillenwater, Kuzman Ganchev, Joao Gra{\c{c}}a, Fernando Pereira, and
  Ben Taskar.
\newblock 2010.
\newblock Sparsity in dependency grammar induction.
\newblock In {\em ACL}.

\bibitem[\protect\citename{Gillenwater \bgroup et al.\egroup
  }2011]{gillenwater2011posterior}
Jennifer Gillenwater, Kuzman Ganchev, Fernando Pereira, Ben Taskar, et~al.
\newblock 2011.
\newblock Posterior sparsity in unsupervised dependency parsing.
\newblock {\em Journal of Machine Learning Research}.

\bibitem[\protect\citename{Gimpel and Smith}2012]{gimpel2012concavity}
Kevin Gimpel and Noah~A Smith.
\newblock 2012.
\newblock Concavity and initialization for unsupervised dependency parsing.
\newblock In {\em NAACL}.

\bibitem[\protect\citename{Grave and Elhadad}2015]{grave2015convex}
Edouard Grave and No{\'e}mie Elhadad.
\newblock 2015.
\newblock A convex and feature-rich discriminative approach to dependency
  grammar induction.
\newblock In {\em ACL-IJCNLP}.

\bibitem[\protect\citename{Han \bgroup et al.\egroup }2017]{han2017dependency}
Wenjuan Han, Yong Jiang, and Kewei Tu.
\newblock 2017.
\newblock Dependency grammar induction with neural lexicalization and big
  training data.
\newblock In {\em EMNLP}.

\bibitem[\protect\citename{Han \bgroup et al.\egroup }2019a]{han2019enhancing}
Wenjuan Han, Yong Jiang, and Kewei Tu.
\newblock 2019a.
\newblock Enhancing unsupervised generative dependency parser with contextual
  information.
\newblock In {\em ACL}.

\bibitem[\protect\citename{Han \bgroup et al.\egroup
  }2019b]{han-etal-2019-multilingual}
Wenjuan Han, Ge~Wang, Yong Jiang, and Kewei Tu.
\newblock 2019b.
\newblock Multilingual grammar induction with continuous language
  identification.
\newblock In {\em EMNLP}.

\bibitem[\protect\citename{He \bgroup et al.\egroup }2018]{he2018unsupervised}
Junxian He, Graham Neubig, and Taylor Berg-Kirkpatrick.
\newblock 2018.
\newblock Unsupervised learning of syntactic structure with invertible neural
  projections.
\newblock In {\em EMNLP}.

\bibitem[\protect\citename{He \bgroup et al.\egroup }2019]{he2019cross}
Junxian He, Zhisong Zhang, Taylor Berg-Kirkpatrick, and Graham Neubig.
\newblock 2019.
\newblock Cross-lingual syntactic transfer through unsupervised adaptation of
  invertible projections.
\newblock In {\em ACL}.

\bibitem[\protect\citename{Headden~III \bgroup et al.\egroup
  }2009]{headden2009improving}
William~P Headden~III, Mark Johnson, and David McClosky.
\newblock 2009.
\newblock Improving unsupervised dependency parsing with richer contexts and
  smoothing.
\newblock In {\em NAACL}.

\bibitem[\protect\citename{Jang \bgroup et al.\egroup
  }2017]{jang2016categorical}
Eric Jang, Shixiang Gu, and Ben Poole.
\newblock 2017.
\newblock Categorical reparametrization with gumble-softmax.
\newblock In {\em ICLR}.

\bibitem[\protect\citename{Jiang \bgroup et al.\egroup
  }2016]{jiang-han-tu:2016:EMNLP2016}
Yong Jiang, Wenjuan Han, and Kewei Tu.
\newblock 2016.
\newblock Unsupervised neural dependency parsing.
\newblock In {\em EMNLP}.

\bibitem[\protect\citename{Jiang \bgroup et al.\egroup
  }2017]{jiang-etal-2017-combining}
Yong Jiang, Wenjuan Han, and Kewei Tu.
\newblock 2017.
\newblock Combining generative and discriminative approaches to unsupervised
  dependency parsing via dual decomposition.
\newblock In {\em EMNLP}.

\bibitem[\protect\citename{Jiang \bgroup et al.\egroup
  }2019]{jiang-etal-2019-regularization}
Yong Jiang, Wenjuan Han, and Kewei Tu.
\newblock 2019.
\newblock A regularization-based framework for bilingual grammar induction.
\newblock In {\em EMNLP-IJCNLP}.

\bibitem[\protect\citename{Kim \bgroup et al.\egroup }2017]{kim2017structured}
Yoon Kim, Carl Denton, Luong Hoang, and Alexander~M Rush.
\newblock 2017.
\newblock Structured attention networks.
\newblock In {\em ICLR}.

\bibitem[\protect\citename{Klein and
  Manning}2004]{Klein:2004:CIS:1218955.1219016}
Dan Klein and Christopher~D. Manning.
\newblock 2004.
\newblock Corpus-based induction of syntactic structure: Models of dependency
  and constituency.
\newblock In {\em ACL}.

\bibitem[\protect\citename{Le and Zuidema}2015]{le2015unsupervised}
Phong Le and Willem Zuidema.
\newblock 2015.
\newblock Unsupervised dependency parsing: Let's use supervised parsers.
\newblock In {\em NAACL}.

\bibitem[\protect\citename{Li \bgroup et al.\egroup }2019]{li2019dependency}
Bowen Li, Jianpeng Cheng, Yang Liu, and Frank Keller.
\newblock 2019.
\newblock Dependency grammar induction with a neural variational
  transition-based parser.
\newblock In {\em AAAI}.

\bibitem[\protect\citename{Li \bgroup et al.\egroup
  }2020]{li-etal-2020-empirical}
Jun Li, Yifan Cao, Jiong Cai, Yong Jiang, and Kewei Tu.
\newblock 2020.
\newblock An empirical comparison of unsupervised constituency parsing methods.
\newblock In {\em ACL}.

\bibitem[\protect\citename{Liu \bgroup et al.\egroup }2013]{liu2013bilingually}
Kai Liu, Yajuan L{\"u}, Wenbin Jiang, and Qun Liu.
\newblock 2013.
\newblock Bilingually-guided monolingual dependency grammar induction.
\newblock In {\em ACL}.

\bibitem[\protect\citename{Ma and Xia}2014]{ma2014unsupervised}
Xuezhe Ma and Fei Xia.
\newblock 2014.
\newblock Unsupervised dependency parsing with transferring distribution via
  parallel guidance and entropy regularization.
\newblock In {\em ACL}.

\bibitem[\protect\citename{Mare{\v{c}}ek and Straka}2013]{marecek2013stop}
David Mare{\v{c}}ek and Milan Straka.
\newblock 2013.
\newblock Stop-probability estimates computed on a large corpus improve
  unsupervised dependency parsing.
\newblock In {\em ACL}.

\bibitem[\protect\citename{Mare{\v{c}}ek and
  {\v{Z}}abokrtsk{\`y}}2012]{marevcek2012exploiting}
David Mare{\v{c}}ek and Zden{\v{e}}k {\v{Z}}abokrtsk{\`y}.
\newblock 2012.
\newblock Exploiting reducibility in unsupervised dependency parsing.
\newblock In {\em EMNLP-IJCNLP}.

\bibitem[\protect\citename{McDonald \bgroup et al.\egroup
  }2005]{mcdonald2005non}
Ryan McDonald, Fernando Pereira, Kiril Ribarov, and Jan Haji{\v{c}}.
\newblock 2005.
\newblock Non-projective dependency parsing using spanning tree algorithms.
\newblock In {\em EMNLP}.

\bibitem[\protect\citename{McDonald \bgroup et al.\egroup
  }2011]{mcdonald2011multi}
Ryan McDonald, Slav Petrov, and Keith Hall.
\newblock 2011.
\newblock Multi-source transfer of delexicalized dependency parsers.
\newblock In {\em EMNLP}.

\bibitem[\protect\citename{Naseem \bgroup et al.\egroup }2010]{naseem2010using}
Tahira Naseem, Harr Chen, Regina Barzilay, and Mark Johnson.
\newblock 2010.
\newblock Using universal linguistic knowledge to guide grammar induction.
\newblock In {\em EMNLP}.

\bibitem[\protect\citename{Nishida and Nakayama}2020]{nishida2020unsupervised}
Noriki Nishida and Hideki Nakayama.
\newblock 2020.
\newblock Unsupervised discourse constituency parsing using {Viterbi EM}.
\newblock {\em TACL}, 8:215--230.

\bibitem[\protect\citename{Noji \bgroup et al.\egroup }2016]{noji2016using}
Hiroshi Noji, Yusuke Miyao, and Mark Johnson.
\newblock 2016.
\newblock Using left-corner parsing to encode universal structural constraints
  in grammar induction.
\newblock In {\em EMNLP}.

\bibitem[\protect\citename{Paskin}2002]{paskin2002grammatical}
Mark~A Paskin.
\newblock 2002.
\newblock Grammatical bigrams.
\newblock In {\em NIPS}.

\bibitem[\protect\citename{Pate and Johnson}2016]{pategrammar}
John~K Pate and Mark Johnson.
\newblock 2016.
\newblock Grammar induction from (lots of) words alone.
\newblock In {\em COLING}.

\bibitem[\protect\citename{Peters \bgroup et al.\egroup }2018]{Peters:2018}
Matthew~E. Peters, Mark Neumann, Mohit Iyyer, Matt Gardner, Christopher Clark,
  Kenton Lee, and Luke Zettlemoyer.
\newblock 2018.
\newblock Deep contextualized word representations.
\newblock In {\em NAACL}.

\bibitem[\protect\citename{Radford \bgroup et al.\egroup
  }2019]{radford2019language}
Alec Radford, Jeffrey Wu, Rewon Child, David Luan, Dario Amodei, and Ilya
  Sutskever.
\newblock 2019.
\newblock Language models are unsupervised multitask learners.
\newblock {\em OpenAI Blog}, 1(8).

\bibitem[\protect\citename{Smith and Eisner}2006]{smith2006annealing}
Noah~A Smith and Jason Eisner.
\newblock 2006.
\newblock Annealing structural bias in multilingual weighted grammar induction.
\newblock In {\em ACL}.

\bibitem[\protect\citename{S{\o}gaard}2012]{sogaard2012unsupervised}
Anders S{\o}gaard.
\newblock 2012.
\newblock Unsupervised dependency parsing without training.
\newblock {\em Natural Language Engineering}, 18(2):187--203.

\bibitem[\protect\citename{Spitkovsky \bgroup et al.\egroup
  }2010a]{SpitkovskyEtAl10}
Valentin~I. Spitkovsky, Hiyan Alshawi, and Daniel Jurafsky.
\newblock 2010a.
\newblock From baby steps to leapfrog: How ``less is more'' in unsupervised
  dependency parsing.
\newblock In {\em NAACL}.

\bibitem[\protect\citename{Spitkovsky \bgroup et al.\egroup
  }2010b]{spitkovsky2010viterbi}
Valentin~I Spitkovsky, Hiyan Alshawi, Daniel Jurafsky, and Christopher~D
  Manning.
\newblock 2010b.
\newblock Viterbi training improves unsupervised dependency parsing.
\newblock In {\em CoNLL}.

\bibitem[\protect\citename{Spitkovsky \bgroup et al.\egroup
  }2011a]{spitkovsky2011unsupervised}
Valentin~I Spitkovsky, Hiyan Alshawi, Angel~X Chang, and Daniel Jurafsky.
\newblock 2011a.
\newblock Unsupervised dependency parsing without gold part-of-speech tags.
\newblock In {\em EMNLP}.

\bibitem[\protect\citename{Spitkovsky \bgroup et al.\egroup
  }2011b]{spitkovsky2011punctuation}
Valentin~I Spitkovsky, Hiyan Alshawi, and Dan Jurafsky.
\newblock 2011b.
\newblock Punctuation: Making a point in unsupervised dependency parsing.
\newblock In {\em CoNLL}.

\bibitem[\protect\citename{Spitkovsky \bgroup et al.\egroup
  }2011c]{spitkovsky2011lateen}
Valentin~I Spitkovsky, Hiyan Alshawi, and Daniel Jurafsky.
\newblock 2011c.
\newblock Lateen {EM:} unsupervised training with multiple objectives, applied
  to dependency grammar induction.
\newblock In {\em EMNLP}.

\bibitem[\protect\citename{Spitkovsky \bgroup et al.\egroup
  }2012]{spitkovsky2012three}
Valentin~I Spitkovsky, Hiyan Alshawi, and Daniel Jurafsky.
\newblock 2012.
\newblock Three dependency-and-boundary models for grammar induction.
\newblock In {\em EMNLP-CoNLL}.

\bibitem[\protect\citename{Spitkovsky \bgroup et al.\egroup
  }2013]{spitkovsky2013breaking}
Valentin~I Spitkovsky, Hiyan Alshawi, and Daniel Jurafsky.
\newblock 2013.
\newblock Breaking out of local optima with count transforms and model
  recombination: A study in grammar induction.
\newblock In {\em EMNLP}.

\bibitem[\protect\citename{Tu and Honavar}2011]{tu2011utility}
Kewei Tu and Vasant Honavar.
\newblock 2011.
\newblock On the utility of curricula in unsupervised learning of probabilistic
  grammars.
\newblock In {\em IJCAI}.

\bibitem[\protect\citename{Tu and Honavar}2012]{tu2012unambiguity}
Kewei Tu and Vasant Honavar.
\newblock 2012.
\newblock Unambiguity regularization for unsupervised learning of probabilistic
  grammars.
\newblock In {\em EMNLP-CoNLL}.

\bibitem[\protect\citename{Yogatama \bgroup et al.\egroup
  }2016]{yogatama2016learning}
Dani Yogatama, Phil Blunsom, Chris Dyer, Edward Grefenstette, and Wang Ling.
\newblock 2016.
\newblock Learning to compose words into sentences with reinforcement learning.
\newblock In {\em ICLR}.

\bibitem[\protect\citename{Yu \bgroup et al.\egroup }2015]{yu2015domain}
Juntao Yu, Mohab El-karef, and Bernd Bohnet.
\newblock 2015.
\newblock Domain adaptation for dependency parsing via self-training.
\newblock In {\em the 14th International Conference on Parsing Technologies}.

\bibitem[\protect\citename{Zhou and Zhao}2019]{zhou2019head}
Junru Zhou and Hai Zhao.
\newblock 2019.
\newblock Head-driven phrase structure grammar parsing on {Penn Treebank}.
\newblock In {\em ACL}.

\end{thebibliography}

\end{document}